%% file: root.tex
\documentclass[letterpaper, 10 pt, journal, twoside]{IEEEtran}
\usepackage{times}

\let\labelindent\relax
\usepackage[numbers]{natbib}
\usepackage{multicol}
\usepackage[bookmarks=true]{hyperref}
\usepackage{microtype}
\usepackage{graphicx}
\usepackage{subfigure}
\usepackage{booktabs} 
\usepackage{etoolbox}
\makeatletter
\patchcmd{\@makecaption}
  {\scshape}
  {}
  {}
  {}
\makeatother
\usepackage[singlelinecheck=false]{caption}
\usepackage{bm}
\usepackage{amsmath}
\usepackage{mathtools}
\usepackage{xcolor}
\usepackage{centernot}

\usepackage{algpseudocode}
\usepackage{centernot}
\usepackage{amsfonts}
\input{defs}

\usepackage[ruled,vlined,commentsnumbered,titlenotnumbered]{algorithm2e}
\usepackage{mdframed}
\begin{document}

\title{
Real-time Model Predictive Control and System Identification Using Differentiable Simulation}


\author{Sirui Chen$^1$, Keenon Werling$^2$, Albert Wu$^2$, C. Karen Liu$^2$
\thanks{Manuscript received: June 29, 2022; Revised: Oct 4; Accepted: Nov 3, 2022.}
\thanks{This paper was recommended for publication by
Editor Lucia Pallottino upon evaluation of the Associate Editor and Reviewers’
comments.}
\thanks{This work is supported by NSF-NRI-2024247, NSF-FRR-2153854, and Stanford-HAI-203112.}
\thanks{$^1$S. Chen is with University of Hong Kong. 
    {\tt\footnotesize ericcsr@hku.hk}}
\thanks{$^2$K. Werling, Albert Wu, C.K. Liu is with Stanford University. 
    {\tt\footnotesize keenon@stanford.edu , karenliu@cs.stanford.edu}}
\thanks{Digital Object Identifier (DOI): see top of this page.}
}
\maketitle
\begin{abstract}
 Transferring a controller from a simulated environment to a physical system is regarded as a challenging problem in robotics. We present a method for continuous improvement of modeling and control \textit{after} deploying the robot to a dynamically-changing target environment. We develop a differentiable physics simulation framework that simultaneously performs online system identification and optimal control using the incoming observations from the target environment in real time. To ensure robust system identification against noisy observations, we devise an algorithm to assess the confidence of our estimated parameters using numerical analysis of the dynamic equations. To ensure real-time optimal control, we adapt start time of the optimization window so that the optimized actions can be replenished ahead of consumption, while staying as up-to-date with new information as possible. The constantly re-planning based on a constantly improving model allows the robot to swiftly adapt to the changing environment using real-world data in a sample-efficient way. Thanks to a fast differentiable physics simulator, both system identification and control can be solved efficiently in real time. We demonstrate our method on a set of examples in simulation and on a real robot. Our method can outperform all baseline methods in different experiments.

\end{abstract}

\IEEEpeerreviewmaketitle

\input{introduction}

\input{related_work}

\input{method}
\input{evaluation}
\input{conclusion}

{\small
\bibliographystyle{IEEEtran}
\bibliography{references}
}
\end{document}

%% file: defs.tex
\usepackage{enumitem}
\setlist[itemize]{itemsep=0pt, topsep=0pt, leftmargin=25pt}
\setlist[enumerate]{itemsep=0pt, topsep=0pt, leftmargin=14pt}

\long\def\ignorethis#1{}

\makeatletter
\DeclareRobustCommand\onedot{\futurelet\@let@token\@onedot}
\def\@onedot{\ifx\@let@token.\else.\null\fi\xspace}

\def\ie{i.e\onedot}

\def\etal{et al\onedot}
\makeatother
\def\shortcite{\def\citename##1{}\@internalcite}

\renewcommand{\eqref}[1]{Equation~(\ref{eq:#1})}

\newcommand{\argmin}{\operatornamewithlimits{argmin}}



%% file: introduction.tex
\section{Introduction}
 Simulation provides a risk-free sandbox for roboticists to develop mechanical designs and control stacks. The major caveat, however, is that using simulation in robotics often leads to control policies or mechanical designs that fail when deployed in the real world,namely the \emph{sim-to-real gap}. Much existing work has focused on training more robust and adaptive control policies in simulation, \emph{prior to} deployment on hardware \cite{sim2real}. In contrast,to enable robot work in constantly changing environment, our work focuses on continuous improvement of modeling and control in the real world, \emph{after} deploying the robot to the dynamically-changing target environment. For example, a quadruped must continuously adapt to terrains with different materials after deployment, and a manipulator must re-estimate the weight of a new object every time for precision control.

\begin{figure}[t]
  \centering
  \includegraphics[width=\linewidth]{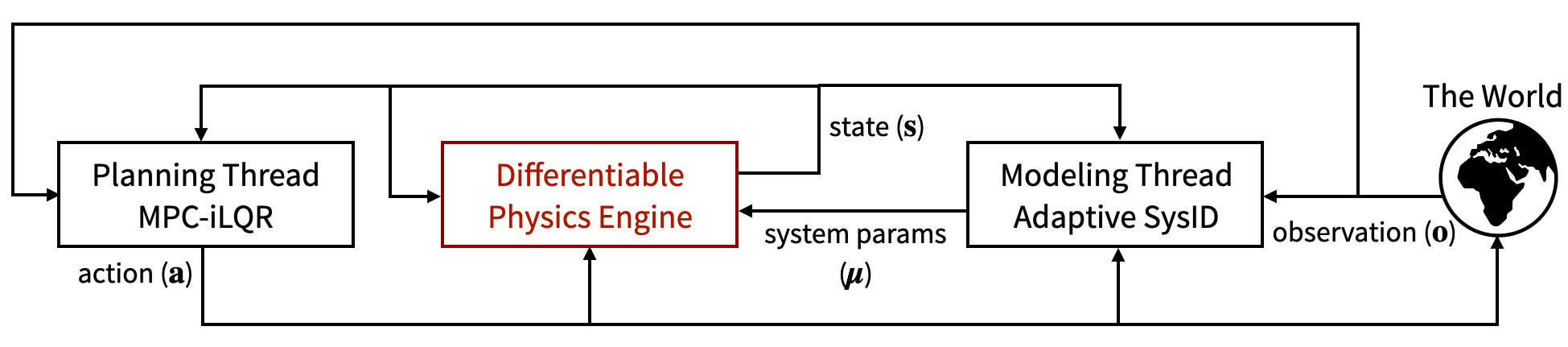}
  \caption{A differentiable physics simulation framework that performs online motion planning and system identification simultaneously.}
  \label{fig:pipeline}
  \vspace{-20pt}
\end{figure}

In this work, we address the problem of post-deployment fine-tuning from both modeling and control perspectives. We develop a differentiable physics simulation framework (Fig.~\ref{fig:pipeline}) that performs online motion planning and system identification simultaneously using incoming observations in real time. Our system runs two parallel threads: a planning thread that solves for a sequence of actions over a finite horizon of time into the \emph{future}, and a modeling thread that optimizes the system parameters based on the most recent \emph{history} of observations. The constant re-planning based on a constantly improving model allows the robot to swiftly adapt to the changing environment and utilize the real-world data in the most sample-efficient way.

The main challenge in online modelling of a dynamically changing environment is that the observations might be outdated (when the environment actually changed) or uninformative (the observations were not “parametrically exciting”). These two scenarios lead to a similar consequence—unstable and erroneous parameters estimation---but call for different solutions. We develop an adaptive online system identification method that monitors the changes in the environment and assesses the confidence of our estimated parameters, using numerical analysis of the dynamic equations. In addition, when the model is inaccurate, our dual-threaded framework provides a unique opportunity to actively control a trajectory for system identification. That is, we can assist the system identification by planning a trajectory that will produce parametrically exciting observations. 

Meanwhile, the main technical challenge for the planning thread is to replenish the planned action buffer faster than real time while keeping the plan as fresh as possible. This is an interesting scheduling problem that can be solved by careful considerations of planning algorithms and problem formulation. We devise a procedure to solve the scheduling problem that balances the accuracy of the model and the computation complexity of the plan. In addition, we leverage the incremental re-planning nature of model-predictive-control to improve the computational performance of the control algorithm.

We evaluate our method on a number of examples both in simulation and on a real robot to demonstrate the effectiveness of our framework for planning and modeling. We compare our method with four baseline methods and evaluate them by the accuracy of system identification and effectiveness of control. Our method outperform all baselines methods.

%% file: related_work.tex
\section{Related Work}
Prior work has explored two approaches to closing the sim-to-real gap: 1) Improving the accuracy and fidelity of robot models and simulators 2) Making the control policy more robust and adaptive to noisy or changing system parameters.

\subsection{Improve modeling accuracy}
System identification (SysID) for robotic systems has been an important research area for decades, but little existing work has proposed effective methods for identifying any parameter of a \emph{generic} physics engine in an \emph{online} fashion. Classical methods, including the frequency and impulse response methods \cite{sysid_survey}, probe the system parameters offline using characteristic input signals. System parameters can also be identified by solving a regression problem for both linear systems \cite{RLS} and nonlinear systems \cite{DOOMED,NLRLS}. Offline SysID can be achieved by differentiable physics engine \cite{DiffCloth,LCP-Physics, GEO}, as well as learned DNNs that approximate dynamic systems \cite{DLN}. However, these methods cannot handle parameter change during control. For online SysID using incoming observations, recursive least square methods have been used for linear systems \cite{RLS}. Moreover, \cite{DUST, EMPPI} solve and update system parameters using stochastic optimization. Recent learning-based approaches attempted to improve modeling accuracy and model unknown time-varying components. For example, Yu \etal \cite{UP-OSI} learned a neural network to model time varying parameters. Jiang \etal \cite{SimGAN} formulated system parameters as functions of state and action which can be learned from real-world trajectories. In contrast, \cite{GPSysID} learns a transition function directly from data without utilizing differential equations. Our adaptive system identification method is designed to run online and handle noisy observations. Unlike learning-based methods, our online SysID does not require offline training using a lot of data.


\subsection{Improve controller design} 

Designing more robust controllers is an effective way to tackle the sim-to-real problem. Robust control aims to battle inaccurate dynamics and noisy measurement. Common Lyapunov functions \cite{CommonLyapunov} can be used to verify controller's robustness, but finding a suitable Lyapunov function for complex nonlinear systems remains a challenge in robotics. Recently, domain and dynamic randomization methods have been used to transfer policies from simulation to different target environments \cite{DBLP:conf/rss/TanZCIBHBV18,DBLP:conf/icra/PengAZA18,DBLP:journals/corr/abs-1901-08652,DBLP:journals/corr/abs-2011-01891}. However, controllers designed to be robust for different parameters often sacrifice task performance.

Adaptive control, on the other hand, relies on an online system identification module to inform the controller with updated parameters. For example, \cite{DUST, EMPPI} estimate the system parameters using stochastic optimization and updated stochastic model predictive controller in an online fashion. \cite{UP-OSI, RMA} used a trained neural network to infer system parameters from historical states, and feed those parameters to a universal policy trained using reinforcement learning. Our method falls in the ``adaptive control'' category, and it demonstrate superior performance compared with previous learning based or stochastic optimization based methods.

\subsection{Differentiable physics}
In the last few years, various differentiable physics simulators have been proposed by researchers aiming to build better simulation tools for robotic tasks such as control design and parameter identification. Prior work has proven differentiable simulation to be a powerful tool for identifying simulation parameters \cite{degrave2019differentiable,difftaichi,Zhong2021NIPS,DiSECT,Wang2021IROS,Lidec2021ICRA}, optimizing design parameters \cite{geilinger2020add,Xu2021RSS,Lutter2021ICRA}, or learn a hybrid simulator which combine learnable neural networks with differential equations \cite{heiden2020augmenting,ContactNets}. The gradient information provided by differential simulation can also be exploited in motion optimization \cite{nimble,DiSECT} and policy learning \cite{Zamora2021ICML}. The unique aspect of our method is that we utilize the gradients provided by a differentiable physics simulator simultaneously for control and system identification in an online fashion. This dual-purpose framework allows a robot to continue to adapt to an ever-changing environment in a sample efficient manner.

%% file: method.tex
\section{Method}
We propose a new method that utilizes differentiable physics for online system identification(SysID) and optimal control simultaneously using streaming observations from the target environment. Our framework runs two parallel threads in real-time, a \emph{planning thread} for optimal control and a \emph{modeling thread} for system identification. The planning thread solves for a sequence of actions over a finite horizon of time into the \emph{future}. Concurrently, the modeling thread optimizes the system parameters, $\bm{\mu}$,  based on the most recent observations. 

\begin{figure*}[t]
  \centering
  \includegraphics[width=\linewidth]{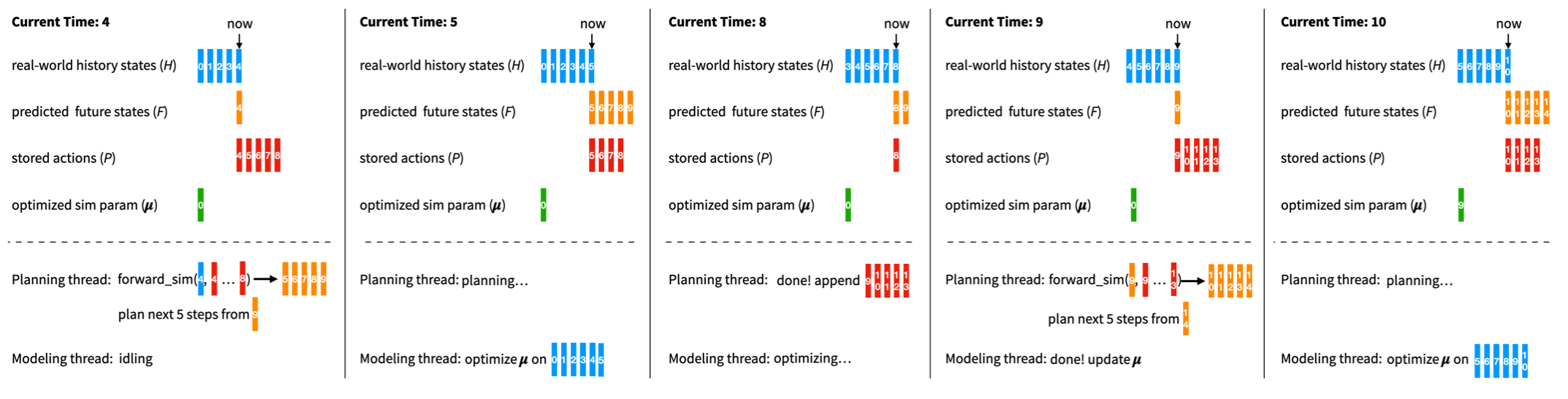}
  \caption{Illustration of our dual-threaded real-time system. At the beginning of time step 4, the history buffer has accumulated five past state, $\bm{x}_{0:4}$ (blue vectors) and the stored plan buffer has 5 actions $\bm{u}_{4:8}$ (red vectors). The system parameters $\bm{\mu}$ were last updated at time step 0 (green vector). Assuming the starting time step and the length of planning horizon is $\hat{t}=5$ and $H=5$ respectively. The planning thread first simulates 5 time steps forward (orange vectors) and starts a trajectory optimization from $\bm{x}_9$. At time step 5, the modeling thread starts optimizing $\bm{\mu}$. At time step 8, the planning thread is done with the new actions $\bm{u}_{9:13}$. At time step 9, the robot starts to consume the new plan while a new trajectory optimization is spawned by the planning thread. At time step 10, the modeling thread updates $\bm{\mu}$.}
  \label{fig:online}
  \vspace{-20pt}
\end{figure*}

Figure \ref{fig:online} illustrates the communication and scheduling between the two threads. The modeling thread takes the state sequence in the history buffer $\mathcal{H}$, which stores the most recent observed states from the target environment, optimizes $\bm{\mu}$ to match $\mathcal{H}$ via the gradients provided by the differentiable physics engine, and finally inform the controller with the optimal $\bm{\mu}$. As soon as the modeling thread completes the optimization, it fetches the next batch of states from $\mathcal{H}$ and repeats the optimization over $\bm{\mu}$. Asynchronously, the planning thread solves for a future action sequence using the differentiable physics engine and the most recent $\bm{\mu}$. The planning thread formulates an optimal control problem as an iterative linear quadratic regulator (iLQR) and solves it in a model-predictive-control (MPC) fashion. The solved action trajectory is placed in the plan buffer $\mathcal{P}$ for the robot to consume in real-time. The next optimal control problem starts immediately after the current one is done with an optimization horizon that extend into the future. 

\begin{algorithm}[b!]
\caption{Online SysID and Control}
\KwIn{$\mathcal{H}$, $\mathcal{P}$} 
$f \longleftarrow f_{task}$
$\bar{\bm{\mu}} \longleftarrow \mathrm{default\_system\_parameters}()$ 
$\bar{w} = 0; \; n = 0$; \\
\Comment{Planning Thread:}\\
\While {$\mathbf{true}$} {
    $t \longleftarrow \mathrm{current\_time}()$ \\
    $\hat{t} = t + t_e \cdot N_{iter}$ \\
    $\bm{x}_{t:\hat{t}} \longleftarrow \mathrm{sim}(\bm{x}_{t},\bm{u_{t:\hat{t}}}; \bar{\bm{\mu}}, \hat{t}-t)$ \\
    $\bm{u}_{\hat{t}:\hat{t}+T-1} \longleftarrow \mathrm{iLQR}(\bm{x}_{\hat{t}}; \bar{\bm{\mu}}, T, f)$ \\
    $\mathcal{P} \longleftarrow \mathrm{update}(\mathcal{P}, \bm{u}_{\hat{t}:\hat{t}+T-1})$ \\
} 
\Comment{Modeling Thread:}\\
\While {$\mathbf{true}$} {
    $\bm{x}_{0:H} \longleftarrow \mathrm{recent\_batch}(\mathcal{H})$ \\
    $\bm{\mu}^* = \argmin_{\bm{\mu}} \| \mathrm{sim}(\bm{x}_0,\bm{u_\text{0:H}}; \bm{\mu}, H) - \bm{x}_{0:H}\|$ \\
    $w^* \longleftarrow \mathrm{confidence\_score}(\bm{x}_{0:H})$ \\
    \If {$|w^*| > \epsilon_1$} {
        $f \longleftarrow t_{task}$ \\
        \If {$\|\bm{\mu}^* - \bar{\bm{\mu}}\| < \epsilon_2$} {
        $\bar{\bm{\mu}} = \bar{\bm{\mu}} \frac{\bar{w}}{\bar{w}+w^*} + \bm{\mu}^* \frac{w^*}{\bar{w} + w^*}$ ,$\bar{w} = \bar{w} + w^*$ \\
        }
        \Else{ 
            $\bar{\bm{\mu}} = \bm{\mu}^*$, $\bar{w} = w^*$
        }
    }
    \Else{
        $n \longleftarrow n+1$ \\
        \If{$n > N_{his}$}{
            $f \longleftarrow f_{exp}$, $n = 0$ \\
        }
    }
    
}

\label{alg:framework}
\end{algorithm} 

\subsection{Real-time system identification}
Given a sequence of recently observed states containing position and velocity in generalized coordinate, $\bm{x}_{0:H} = [\bm{q}_{0:H},\dot{\bm{q}}_{0:H}]$, in the history buffer $\mathcal{H}$, a standard SysID routine finds the optimal system parameters that best fit the observations: $\boldsymbol{\mu}^* = \arg\min_{\bm{\mu}}||\mathrm{sim}(\bm{q}_0,\dot{\bm{q}}_0,\bm{u_\text{0:H}}; \bm{\mu}, H) - \bm{q}_{0:H}||$, where $\mathrm{sim}(\bm{q}_0,\dot{\bm{q}}_0, \bm{u_\text{0:H}}; \bm{\mu}, H)$ is the forward simulation starting from state $(\bm{q}_0,\dot{\bm{q}}_0)$ for $H$ time steps \textcolor{black}{controlled by $\bm{u_\text{0:H}}$} under the differential equations parameterized by $\bm{\mu}$. Utilizing a differential physics simulator (e.g. NimblePhysics \cite{nimble}), we can compute the gradients of the objective function efficiently to optimize $\bm{\mu}$ in real time. 

However, unlike a one-time offline SysID routine, repeatedly applying online SysID using incoming noisy observations in a dynamically changing environment often leads to erroneous and unstable estimation of system parameters. The main challenge of online SysID is that the observations might be outdated (when the robot model actually changed) or uninformative (the observations were not “parametrically exciting”). The former requires the system to flush old observations in the buffer and only use the recent observations, while the latter can be mitigated by weighing the old observations more highly than the recent ones. Knowing which failure mode we are currently in is critical for the system to respond correctly.

We develop an adaptive online SysID method, inspired by the idea of persistent excitation analysis\cite{persistent}, to assess how a state trajectory spans the range of system behaviors. For each SysID optimization problem, we compute a confidence score, $w$ and weight the solution by the confidence score. The currently estimated system parameter is defined as the weighted sum of solutions found in the past:
\begin{equation}
\label{eqn:weighted_solution}
\bar{\boldsymbol{\mu}} = \frac{\sum_{t=t_0}^{T_\text{now}}w_t \; \boldsymbol{\mu}_t}{\sum_{t=t_0}^{T_\text{now}}w_t}.
\end{equation}

At the end of each SysID optimization, we analyze the solution $\bm{\mu}^*$ and the confidence score $w^*$ to determine the appropriate action. If $w^*$ is large and $\|\bm{\mu}^* - \bar{\bm{\mu}}\|$ is small (\ie the new solution is similar to the current one), we accept the new solution $\bm{\mu}^*$ and update $\bar{\bm{\mu}}$ using Equation \ref{eqn:weighted_solution}. If $w^*$ and $\|\bm{\mu}^* - \bar{\bm{\mu}}\|$ are both large, it is likely that the environment has changed and $\bm{\mu}$ needs to be re-estimated from scratch. In this case, we replace $\bar{\bm{\mu}}$ with $\bm{\mu}^*$. If $w^*$ is small, we discard $\bm{\mu}^*$ and do not update $\bar{\bm{\mu}}$. If small $w^*$ persists for a $N_{his}$ SysID optimizations, we switch to active SysID mode and change the objective function in the planning thread (details in Section \ref{sec:active_sysID}). Algorithm \ref{alg:framework} summarizes both the adaptive online SysID and optimal control.

To compute the confidence score $w$, we use the Lagrange's equations of motion for articulated rigid body systems to estimate each system parameter's excitation:
\begin{equation}
\label{eqn:EOM}
\begin{split}
\bm{M}(\bm{q}; \bm{\mu})\ddot{\bm{q}} + \bm{c}(\bm{q}, \dot{\bm{q}}; \bm{\mu}) + \bm{g}(\bm{q}; \bm{\mu}) =  \bm{f}(\bm{q}, \dot{\bm{q}};\bm{\mu}),
\end{split}
\end{equation}
where $\bm{M}$ is the mass matrix in generalized coordinates $\bm{q}\in\mathbb{R}^N$, $\bm{c}$ is the Coriolis and centrifugal force, $\bm{g}$ is the gravity, and $\bm{f}$ is the sum of other generalized forces applied on the system. While the actual formula to compute $w$ for each system parameter is different, the underlying principle is the same: \textcolor{black}{For linear parameters, we directly analyze excitation based on equation of motion; for nonlinear parameters, we either drop some insignificant nonlinear terms or linearize the system using differentiable physics engine before excitation analysis.} We derive the formulations for four common parameters below.

\paragraph{Confidence score for estimating masses}
To factor out the mass of each link, $m_k$, we rewrite Equation \ref{eqn:EOM} as a linear function of $\bm{m} = [m_1, \cdots, m_M]$, where $M$ is the number of rigid links in the system. For clarity of exposition, we omit the Coriolis and centrifugal force.
\begin{equation}
    \big(\bm{A}(\bm{q}, \ddot{\bm{q}})+\bm{B}(\bm{q})\big) \bm{m} = \bm{f},
\end{equation}
where the column $k$ in $\bm{A} \in\mathbb{R}^{N \times N}$ represents the acceleration of the rigid link $k$ due to the inertial force in the generalized coordinates: $\bm{A}[:,k] = (\bm{J}_k^T \bm{J}_k + \bm{J}_{\omega k}^T \tilde{\bm{I}}_k \bm{J}_{\omega k}) \ddot{\bm{q}} \in \mathbb{R}^{N}$ \footnote{We loosely use the matrix indexing notation $[:,k]$ to indicate the column $k$ of the matrix.}. $\tilde{\bm{I}}_k \in \mathbb{R}^{3 \times 3}$ is the inertia matrix in the coordinate frame of link $k$ with the mass $m_k$ factored out. The Jacobian matrix for the rigid link $k$ contains two parts: the linear Jacobian $\bm{J}_k \in \mathbb{R}^{3 \times N}$ and the angular Jacobian $\bm{J}_{\omega k}\in \mathbb{R}^{3 \times N}$ which together map $\dot{\bm{q}}$ to the linear velocity and the angular velocity of link $k$ in the Cartesian space. Similarly, the column $k$ in $\bm{B}$ represents the acceleration of the rigid link $k$ due to the gravitational force in the generalized coordinates: $\bm{B}[:,k] = \bm{J}_k^T \bm{g}_e \in \mathbb{R}^{N}$, where $\bm{g}_e = (0, 0, -9.8)^T$.

If the observations happen to make the rank of $\bm{A} + \bm{B}$ less than $M$, we cannot uniquely identify the mass for all $M$ links. However, solving the rank for each time step and for each rigid link can be time consuming. We opt to use a heuristic that measures the magnitude of each column in $\bm{A}+\bm{B}$. When  $\|\bm{A}[:,k]+\bm{B}[:,k]\|$ is small, the estimate of $m_k$ is more sensitive to the sensing noise. If $\|\bm{A}[:,k]+\bm{B}[:,k]\|$ is zero, we lose the rank to identify $m_k$ altogether. As such, the confidence score for estimating mass is defined as:
\[
w = \sum_{t=0}^{H-1}\sum_{k=0}^{M-1} \|\frac{1}{\Delta t}(\bm{J}_k^T \bm{J}_k + \bm{J}_{\omega k}^T \tilde{\bm{I}}_k \bm{J}_{\omega k})(\dot{\bm{q}}_{t+1} - \dot{\bm{q}}_t) + \bm{J}_k^T\bm{g}_e\|,
\]
where $\Delta t$ is the simulation time step. \textcolor{black}{Jacobian matrices can be typically obtained from the physics engine directly in practice}

\paragraph{Confidence score for estimating moment of inertia}
Estimating the moment of inertia of each rigid link is also common for system identification. In our confidence score computation, we do not consider the off-diagonal elements in the local inertia matrix $\bm{I}_k$, as they are typically dominated by the diagonal elements (the principle inertia). Rearranging Equation \ref{eqn:EOM} to factor out the principle inertia terms, $\bm{d}_k = (I_k^{xx}, I_k^{yy}, I_k^{zz})$, for each link $k$, we arrive at:
\begin{equation}
    \begin{bmatrix}\bm{C}_1(\bm{q}, \ddot{\bm{q}}) & \cdots & \bm{C}_M(\bm{q}, \ddot{\bm{q}})\end{bmatrix}\begin{bmatrix} \bm{d}_{1} \\ \vdots \\ \bm{d}_{M}\end{bmatrix} = \bm{b},
\label{eqn:estimate_inertia}
\end{equation}
where $\bm{C}_k = \bm{J}^T_{\omega k} \mathrm{Diag}(\bm{J}_{\omega k} \ddot{\bm{q}}) \in \mathbb{R}^{N \times 3}$ and $\bm{b} = \bm{f} - \bm{g} - \sum_{k=0}^{M-1} m_k \bm{J}_k^T \bm{J}_k \ddot{\bm{q}} \in \mathbb{R}^N$, the sum of all the terms in Equation \ref{eqn:EOM} independent of the inertia matrix. The operator Diag($\bm{v}$) maps a vector $\bm{v} \in \mathrm{R}^n$ to a $\mathrm{R}^{n\times n}$ diagonal matrix with $\bm{v}$ as its diagonal elements. Analyzing the rank of the matrix on the LHS of Equation \ref{eqn:estimate_inertia} is too costly. We define a simpler confidence score similar to the one for the mass estimation:
\[
w = \sum_{t=0}^{H-1} \sum_{k=0}^{M-1} \sum_{i=0}^2 \|\frac{1}{\Delta t}\bm{J}^T_{\omega k} \mathrm{Diag}(\bm{J}_{\omega k} (\dot{\bm{q}}_{t+1} - \dot{\bm{q}}_t))(:,i)\|
\]

\paragraph{Confidence score for estimating center of mass}
The center of mass (COM) of each body link $k$ affects the equations of motion via the Jacobian $\bm{J}_k$, which maps the generalized velocity $\dot{\bm{q}}$ to the Cartesian velocity of COM, $\bm{v}_k$. We can compute $\bm{v}_k$ recursively from the Jacobian of its parent joint $\tilde{\bm{J}}_k \in \mathbb{R}^{3 \times N}$:
\begin{equation}
    \bm{v}_k = \tilde{\bm{J}}_k \dot{\bm{q}} + \bm{R}_k[\hat{\bm{J}}_{\omega k}\dot{\bm{q}}]\bm{r}_k = (\tilde{\bm{J}}_k  - \bm{R}_k [\bm{r}_k]\hat{\bm{J}}_{\omega k}) \dot{\bm{q}},
\label{eqn:linear_velocity}
\end{equation}
where $\hat{\bm{J}}_{\omega k} \in \mathbb{R}^{3 \times N}$ maps the generalized velocity to the angular velocity of link $k$ expressed in the frame of parent joint, $\bm{R}_k$ is the transformation from the parent joint frame of link $k$ to the world frame, and $\bm{r}_k$ is the COM in the frame of the parent joint of link $k$. The bracket $[\;]$ indicates the skew symmetric matrix. By the definition of $\bm{J}_{k}$ and Equation \ref{eqn:linear_velocity},
\begin{equation}
\bm{J}_k = \tilde{\bm{J}}_k  - \bm{R}_k [\bm{r}_k]\hat{\bm{J}}_{\omega k}.
\label{eqn:J_k}
\end{equation}

The terms on the LHS of Equation \ref{eqn:EOM} dependent on $\bm{J}_k$ can be expressed as $m_k\bm{J}_k^T(\bm{J}_k\ddot{\bm{q}}+\dot{\bm{J}}_k\dot{\bm{q}}-\bm{g}_e)$. Substituting $\bm{J}_k$ with Equation \ref{eqn:J_k} and dropping the quadratic terms in $\bm{r}_k$ and the constant mass $m_k$ \footnote{Dropping the quadratic terms can affect the rank analysis and is considered part of approximation by our heuristic.}, we can express the linear terms in $\bm{r}_k$ on the LHS of Equation \ref{eqn:EOM} as
\begin{equation}
\begin{split}
    \Big(\tilde{\bm{J}}_k^T\big( \bm{R}_k[\hat{\bm{J}}_{\omega k} \ddot{\bm{q}}] + \dot{\bm{R}}_k [\hat{\bm{J}}_{\omega k} \dot{\bm{q}}] + \bm{R}_k [\dot{\hat{\bm{J}}}_{\omega k} \dot{\bm{q}}]\big) \\ + \hat{\bm{J}}_{\omega k}^T [\bm{R}_k^T (\tilde{\bm{J}}_k \ddot{\bm{q}}  + \dot{\tilde{\bm{J}}}_k \dot{\bm{q}} - \bm{g}_e)]\Big) \bm{r}_k.
\end{split}
\label{eqn:eom_com}
\end{equation}
 
To identify COM $\bm{r}_k$ for all $M$ links, we need to concatenate the matrix expressed in Equation \ref{eqn:eom_com} (inside of the parenthesis) for every link into one single $N \times 3M$ matrix and analyze its rank. Similar to other system parameter estimation, we use a heuristic to approximate the rank. 

Let $\bm{S}$ be $\bm{R}_k[\hat{J}_{\omega k}\ddot{\bm{q}}]+\dot{\bm{R}_k}[\hat{\bm{J}}_{\omega k}\dot{\bm{q}}]+\bm{R}_k[\dot{\hat{\bm{J}}}_{\omega k}\dot{\bm{q}}]$ and $\bm{G}$ be $[\bm{R}_k^T (\tilde{\bm{J}}_k \ddot{\bm{q}} + \dot{\tilde{\bm{J}}}_k \dot{\bm{q}} - \bm{g}_e)]$, both in $\mathbb{R}^{3 \times 3}$. Expression in  \ref{eqn:eom_com} can be simplified to $\tilde{\bm{J}}_k^T \bm{S} + \hat{\bm{J}}_{\omega k}^T \bm{G}$, which needs to be full column-rank for $\bm{r}_k$ to be identifiable. If $\bm{r}_k$ is already in the nullspace of $\bm{S}$, multiplying $\tilde{\bm{J}}_k^T \in \mathbb{R}^{N \times 3}$ is not going to make $\bm{r}_k$ identifiable. The same argument can be made for the $\bm{G}$ term. Therefore, we can devise a simple heuristic based on the necessary (not sufficient) condition for every $\bm{r}_k$ to be identifiable:
\[
 w = \sum_{t=0}^H \sum_{k=0}^M \sum_{i=0}^2 \|\bm{S}_{k} [:,i]\| + \|\bm{G}_{k} [:,i]\|
\]

\paragraph{Confidence score for estimating joint stiffness and damping}
The confidence scores for the joint stiffness and damping are more straightforward. We model the generalized force due to joint stiffness and damping as $\bm{f}_{joint}(\bm{q}, \dot{\bm{q}}; \bm{\mu}) = -\bm{K}_s (\bm{q} - \bar{\bm{q}}) - \bm{K}_d \dot{\bm{q}}$, where $\bar{\bm{q}}$ is the preset rest position for the joint angles. The confidence score for the joint stiffness can be computed by: $
w = \sum_{t=0}^{T-1} \|\bm{q}_t-\bar{\bm{q}}\|$, and the confidence for the damping can be computed by $w = \sum_{t=0}^{T-1} \|\dot{\bm{q}}_t\|$.

\subsection{Real-time optimal control}
The real-time optimal control problem can be formulated either as a nonlinear program, which can be solved with a general optimization package (e.g., IPOPT \cite{IPOPT}), or as an iLQR \cite{iLQR} through leveraging recursive structure of the problem. With an efficient differentiable physics engine, our method can be applied to both control approaches. However, we prefer iLQR as its runtime is more predictable for online control tasks. The computation speed of MPC-iLQR can be significantly improved with warm-starting using the overlapping time window from the previous solution. Specifically, we initialize the linear control laws, the state trajectory, and the action trajectory using the previous solution.

The main technical challenge for the planning thread is to replenish the plan buffer $\mathcal{P}$ faster than real-time while keeping the plan as fresh as possible. We propose to adaptively select the starting time $\hat{t}$ of the horizon $T$ of the trajectories being optimized: $\bm{x}_{\hat{t}:\hat{t}+T}, \bm{u}_{\hat{t}:\hat{t}+T-1}$ where $\bm{x},\bm{u}$ are state and action of the robot. Ideally we would like $\hat{t}$ to be as close as possible to the the time index when the trajectory optimization is finished, so the new plan $\bm{u}_{\hat{t}:\hat{t}+T-1}$ will be fresh and just-in-time. If $\hat{t}$ is too early, some of the new plan will be stale already by the time the planning is done. If $\hat{t}$ is too late, the robot cannot switch to the new plan when the old one is depleted and will lose control afterward.

We profile the runtime of the planning algorithm running on the target hardware to obtain the duration $t_e$ for each iLQR iteration. During runtime, we determine $\hat{t}$ by multiplying the number of iterations, $N_{iter}$, the previous iLQR took with $t_e$: $\hat{t} = t +t_e \cdot N_{iter} $, where $t$ is the current time index (Alg \ref{alg:framework}). 

\subsection{Actively controlled system identification}
\label{sec:active_sysID}
An advantage of our dual-threaded framework is that we can leverage the planning thread to assist the modeling thread when SysID struggles to identify the system parameters for a long time. If the confidence score has been low for $N_{his}$ SysID solutions consecutively, we inform the planning thread to switch its objective function from the main task $f_{task}$ (See \ref{sec:evaluation} for detailed definition) to $f_{exp}$ which is designed to explore the range of system behaviors. Conveniently, the confidence score described above can be reused for actively planning a state trajectory that maximizes the parametric excitation for the system parameters of interest. Therefore, we define $f_{exp} = -w(\bm{x}_{0:H})$ (Algorithm \ref{alg:framework}).

%% file: evaluation.tex
\section{Evaluation}
\label{sec:evaluation}
\begin{table}[t]
    \centering
    \resizebox{0.85\linewidth}{!}{
    \begin{tabular}{@{}lrrrrr@{}}
    \toprule
    \multicolumn{5}{c}{Hyper parameters setting}\\
    \hline
             & Cartpole   & InvDP      & Arm(COM) & Arm(MOI) & Elastic Rod\\
    \hline
    $\epsilon_1$     & 0.02 & 0.05 & 0.02 & 0.02 & 0.02\\
     $\epsilon_2$ & 0.5 & 0.5 & 0.5 & 0.5 & 0.5\\
     $H$   & 5 & 10 & 5 & 5 & 10\\
     $T$ & 100 & 200 & 100 & 100 & 300\\
     Range & [0.2, 5.0] & [0, 0.5] & [0, 0.05] & [0, 0.2] & [0, 15]\\
     \bottomrule
    \end{tabular}
    }
    \caption{Hyperparameters for our experiments. $\epsilon_1$ is the normalized threshold for detecting parameter changes, $\epsilon_2$ is the threshold of confidence, $H$ is the trajectory length for SysID, and $T$ is the horizon for iLQR. The unknown parameters are randomly sampled in the ranges shown here. The range of COM and MOI are determined by geometric boundary of the robot arm.}
    \label{tab:params}
    \vspace{-20pt}
\end{table}

We implemented our method using an off-the-shelf differentiable physics engine, NimblePhysics \cite{nimble}. We evaluate our method on four dynamic motor control tasks in which the robot needs to continuously identify the system parameters and solve for the control trajectories. We compare our methods to four different baselines: 
\begin{enumerate}
    \item {\bf{Naive:}} Solve SysID using the most recent observations. Solve control using MPC-iLQR.
    \item {\bf{Smooth:}} Solve SysID using the average of five previously solved system parameters. Solve control using MPC-iLQR.
    \item {\bf{Weighted:}} Solve SysID using Equation \ref{eqn:weighted_solution} with five previously solved system parameters. Solve control using MPC-iLQR. This baseline is the same as our method except that the most recent solution is always accepted and interpolated with the current solution.
    \item{\bf{UP-OSI:}} Use an MLP model trained offline for SysID. Control the robot using a system-parameter-conditioned policy trained offline using deep RL approach. We use the implementation by \cite{UP-OSI}.
    \item{\bf{DuST:}} Use stochastic optimization for both online SysID and model predictive control \cite{DUST}.
\end{enumerate}

We also evaluate the effectiveness of adaptive horizon starting time for the planning thread. The optimal control problem for all four tasks share the same form of objective functions:
\begin{equation}
\label{eqn:LQR}
\begin{split}
    f_{task}(\bm{x}_{0:T}, \bm{u}_{0:T-1}) = \frac{1}{2}\sum_k^{T-1} (\bm{x}_k-\bar{\bm{x}})^T\bm{Q}_r(\bm{x}_k-\bar{\bm{x}}) \\+ \frac{1}{2}(\bm{x}_{T}-\bar{\bm{x}})^T\bm{Q}_f(\bm{x}_{T}-\bar{\bm{x}}) + \frac{1}{2}\sum_k^{H-1} \bm{u}_k^T\bm{R}\bm{u}_k,
\end{split}
\end{equation}
where $\bar{\bm{x}}$ is the target state, $\bm{R}$ is the cost weights for an action, $\bm{Q}_r$ and $\bm{Q}_f$ defines the running cost and final cost weights for a state. In the four tasks we demonstrated, we found that only one example, the elastic rod moving task, triggers the active system identification mode in which the objective function for iLQR switched to $f_{exp} = -w(\bm{x}_{0:H})$.  

To bring the simulated environments closer to the real-world, we simulate sensing and actuating errors by adding random gaussian noise in the observed states, as well as in the actions being executed. All the hyper parameters are reported in Table \ref{tab:params}.


\subsection{Swing-Up cartpole with unknown masses}
We started with a classical motor control problem: swing up and balance a pole attached to a cart by applying a force to the cart. The target $\bar{\bm{x}}= [1, 0, 0 ,0]$ requires the pole to be balanced at the configuration $(1, 0)$ with zero velocity as quickly as possible. 

\begin{figure}[t]
  \centering
  \includegraphics[width=0.9\linewidth]{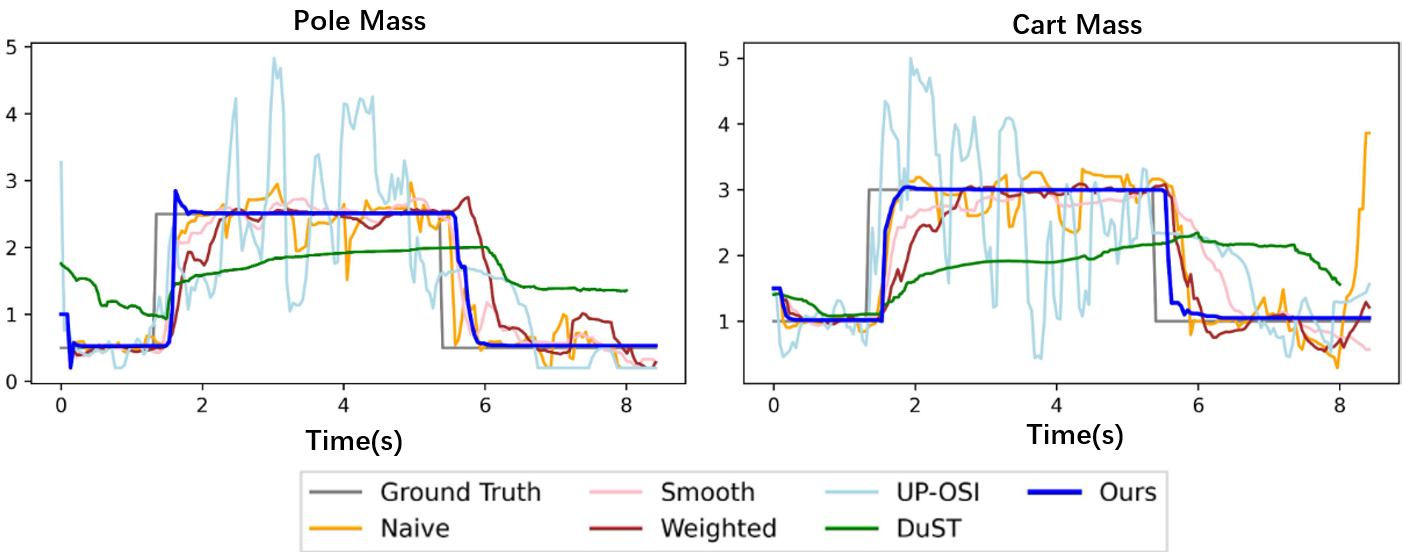}
  \caption{Identifying masses for the cartpole experiment.}
  \label{fig:cartpole}
\end{figure}

The mass of the cart and the mass of the pole are unknown initially and will change twice during control. Figure \ref{fig:cartpole} shows the accuracy and stability of our method compared against the baselines. Our method can quickly identify the correct masses and respond rapidly to the parameter change. We further demonstrated that more effective SysID positively impacts the quality of control. Table \ref{tab:control} shows that the time required for our MPC-iLQR controller to complete the task. Our controller is significantly more effective than baselines based on the same MPC-iLQR. UP-OSI and DuST was not able to complete the task after $1000$ time steps, suggesting that the MLP-based SysID can be brittle when the changes in parameters result in out-of-distribution state trajectories (See the accompanying video for interactive demos). Without utilizing efficient gradient based optimization DuST failed to respond rapidly to system parameter change. It also worth mentioning our controller is 80x faster than the stochastic optimization based controller implemented in DuST. Both comparison with DuST has shown that utilizing a differentiable model can significantly accelerate control optimization and adaptation.




\begin{table}[t]
    \centering
    \resizebox{0.80\linewidth}{!}{
    \begin{tabular}{@{}lrrrr@{}}
    \toprule
    \multicolumn{5}{c}{Time (second) required to reach control objective}\\
    \hline
             & Cartpole   & InvDP      & Robot Arm & Elastic Rod\\
    \hline
    Ours     & \textbf{6.31(0.42)} & \textbf{3.53(0.22)} & \textbf{2.10}(0.20) & \textbf{4.60(0.07)}\\
    \hline
    Weighted & 6.76(0.45) & 3.73(0.46) & 2.53(0.05) & 6.42(1.06)\\
    \hline
    Smooth   & 6.98(0.62) & 3.80(0.44) & 2.76(0.07) & 6.70(2.57)\\
    \hline
    Naive    & 7.57(0.87) & 4.26(0.61) & 3.01(0.10) & 9.05(4.08)\\
    \hline
    UP-OSI   & Failed     & 6.23(0.75) & NA         & NA        \\
    \hline
    DuST     & Failed     & Failed     & NA         & NA\\
    \bottomrule
    \end{tabular}
    }
    \caption{Time required to complete the control task. Mean and standard deviation over 5 experiments with different random seeds are reported. A trial is considered successful if it reaches a state $x$ where $||x-x_\text{target}||_2\leq 0.1$ within $1000$ time steps. We only compared with UP-OSI and DuST on the tasks implemented in their public codebases.}
    \label{tab:control}
    \vspace{-20pt}
\end{table}

\subsection{Inverted double pendulum (InvDP) with unknown damping}
This task is the same as the cart-pole, but an additional unactuated pole segment makes the double inverted pendulum a more challenging control problem.  The target state $\bar{\bm{x}}$ is set to be $[0, 0, 0, 0, 0, 0]$ which requires the double inverted pendulum to be balanced stably at upright configuration as quickly as possible. 

\begin{figure}[t]
  \centering
  \includegraphics[width=0.9\linewidth]{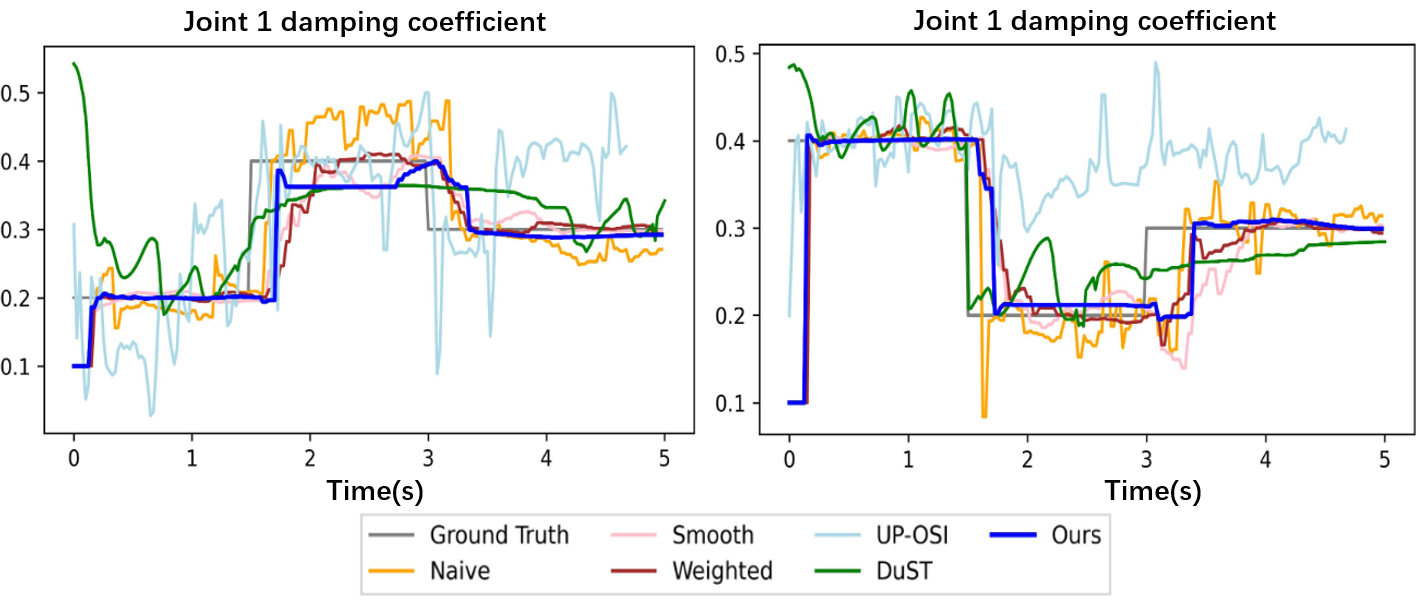}
  \caption{Identifying damping coefficients for the double inverted pendulum experiment. }
  \label{fig:dinvp}
\end{figure}

The damping coefficients of the two pin joints are unknown. Figure \ref{fig:dinvp} shows that these parameters can be more accurately identified by our method compared to others. We further demonstrated that the accuracy of SysID plays a significant role in control. Table \ref{tab:control} shows the time required to complete the task. Our method again outperforms other baselines.

\subsection{Robot arm with unknown end-effector mass distribution}
Robot manipulators often encounter objects with unknown mass and unknown mass distribution. While the mass of an object is simple to obtain from force sensors, estimating the mass distribution of an object is more challenging. Furthermore, for objects like a bag of sand or a jar of water, the center of mass and the moment of inertia can change over time and need to be estimated in an online fashion.

In this example, a Rokae Xmate3P robotic arm is required to carry an object from the initial configuration of the robot to a final configuration. The COM and the moment of inertial of the object is unknown and could change over time. 

\begin{figure}[t]
  \centering
  \includegraphics[width=0.9\linewidth]{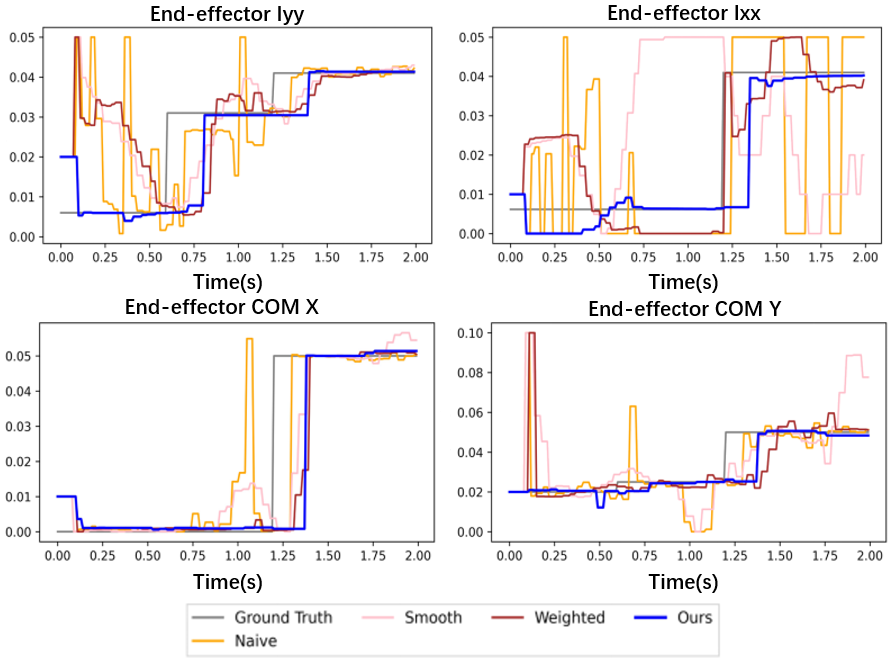}
  \vspace{-5pt}
  \caption{Identifying the center of mass and the moment of inertia for Rokae Xmates3 robot arm. Since the end-effector cannot rotate in the roll direction, we only show the identification result around X and Y axis. }
  \label{fig:arm}
  \vspace{-20pt}
\end{figure}

We assumed that the object is rigidly attached to the end effector of the robot. The center of mass and the moment of inertia of the end effector-object assembly are modified during the experiment. Figure \ref{fig:arm} shows that our adaptive SysID outperforms all the baselines on tracking system parameters. Similarly, the accurate estimation of the system parameters can accelerate control convergence as shown in Tab. \ref{tab:control}.

\subsection{Moving an elastic rod with unknown elasticity}
This example involves an elastic rod modeled by four rigid links connected by revolute joints with internal springs. The task is to push the base link of the elastic rod to the target location as quickly as possible, while keeping the rod straight and vertical when hitting the target. Intuitively, if the rod is stiff, we can apply a relatively large acceleration to the base. If the rod is more elastic, then we have to accelerate strategically to mitigate bending. Accurate estimation of spring stiffness is crucial for faster control.

\begin{figure}[t]
  \centering
  \includegraphics[width=0.9\linewidth]{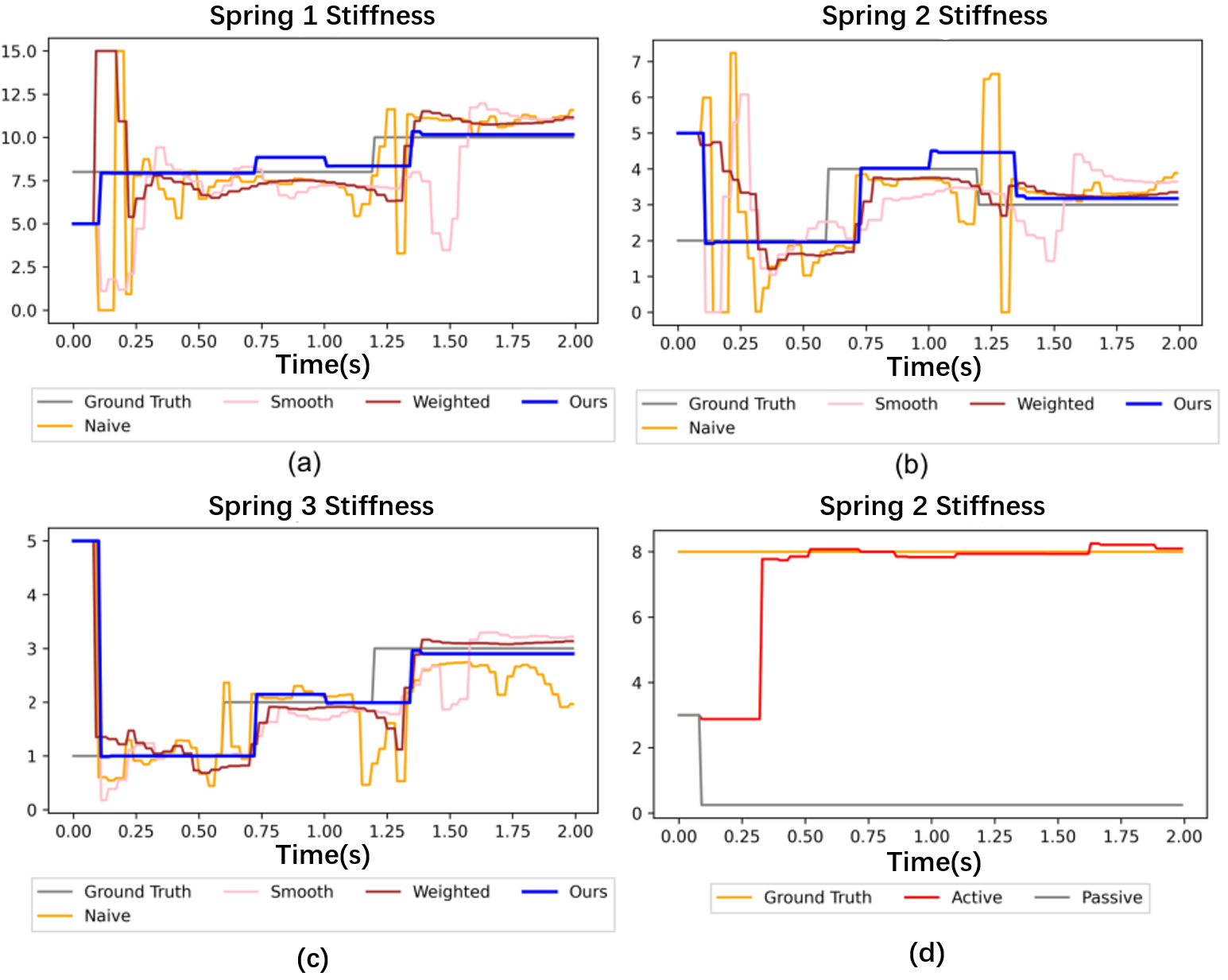}
  \vspace{-5pt}
  \caption{(a)-(c): Identifying the spring stiffness for the elastic rod.(d): System identification w/ and w/o active SysID control.}
  \label{fig:elastic_rod}
  \vspace{-20pt}
\end{figure}

This is an example that benefits from the coordination of the modeling thread and the planning thread, such that correct parameters can be identified and the task can be accomplished efficiently. At the beginning, the iLQR with underestimated joint stiffness parameters produces slow actions in the hope of keeping the rod straight when it reaches the target. However, these conservative actions fail to excite the system and prevent the joint stiffness from being identified. Our confidence score correctly exposed the issue and switched to the actively controlled SysID mode. Fig \ref{fig:elastic_rod} (d) shows that active SysID quickly corrects the joint stiffness and switches back to optimizing the task objective, resulting in a faster control sequence at the end. Active SysID control can further reduce the average time to reach the target by 2.072 seconds.

We compare our results with the baselines and \ref{fig:elastic_rod} show that our method can identify the elasticity of the rod accurately and respond quickly to changing parameters. Notice that the \textbf{Weighted} baseline also outperform the other baselines, showing that our confidence score is effective even with a simple moving average scheme. The result for control is shown in Table \ref{tab:control}

\subsection{Study on adaptive starting time}
We evaluated the impact of adaptive starting time for the planning thread on the cart-pole example. \textcolor{black}{We measure the accuracy of estimation. On average, each optimization problem takes 20.8ms to solve, and MAE between the actual start time and the estimated one is 3.04ms.} Furthermore, we replace our adaptive method with a fixed starting time. To pick a reasonable fixed starting time, we use the average time required per planning. Over five experiments, we found that the fixed starting time on average takes 8.20 seconds to swing up and balance the cartpole with two failed cases, whereas our method on average takes 6.31 seconds without failure.

\subsection{\textcolor{black}{Hyper-parameter selection}}
\textcolor{black}{We use the Cartpole experiment to study the effect of different hyper parameters selection of $\epsilon_1$, $\epsilon_2$, $H$, and $T$. Table \ref{tab:param_search} shows that our model is relatively robust against different selection of SysID horizon $H$, threshold for detecting parameter changes $\epsilon_1$, and confidence threshold $\epsilon_2$. Although it is more  sensitive to the MPC control horizon $T$, it is not uncommon that $T$ requires hand-tuning for different applications in practice.}

\begin{table}[t]
    \centering
    \resizebox{0.9\linewidth}{!}{
    \begin{tabular}{@{}lrlrlrlr@{}}
    \toprule
    \multicolumn{8}{c}{Comparison of Different Hyper-parameters selection}\\
    \hline
    $\epsilon_1$         & $t_\text{ctrl}$   & $\epsilon_2$      & $t_\text{ctrl}$ & H & $t_\text{ctrl}$ & T & $t_\text{ctrl}$\\
    \hline
    $2\times 10^{-3}$ & 6.58(1.0) & 0.1 & 8.57(0.3) & 2 & 11.71(2.0) & 50 & 7.79(5.3)\\
    $2\times 10^{-2}$  & 6.31(0.4) & 0.5 & 6.31(0.4) & 5 & 6.31(0.4)  & 100 & 6.31(0.4)\\
    $2\times 10^{-1}$   & 6.92(3.0) & 0.7 & 8.91(1.0) & 10& 6.32(1.6)  & 150 & 11.01(3.0)\\
    $5\times 10^{-1}$   & 9.00(1.4) & 0.9 & 13.64(5.3)& 20& 7.55(2.8)  & 200 & 14.60(3.4)\\
     \bottomrule
    \end{tabular}
    }
    \caption{\textcolor{black}{The effect of hyper parameters on the control performance of Cartpole. $t_{ctrl}$ indicates the time required for the cartpole to achieve the control goal.}}
    \label{tab:param_search}
\end{table}

\subsection{\textcolor{black}{System ID with contact}}
\begin{figure}[t]
  \centering
  \includegraphics[width=0.7\linewidth]{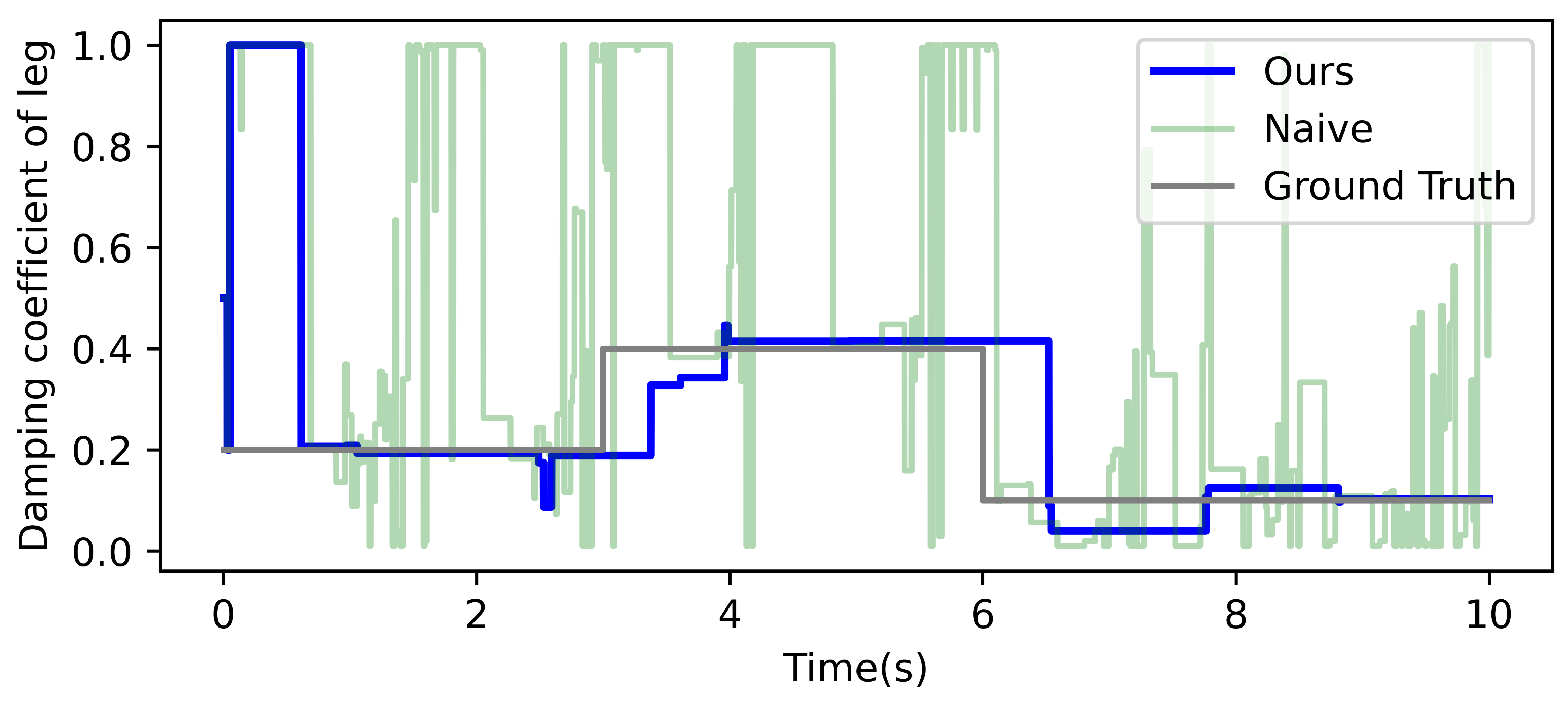}
  \vspace{-5pt}
  \caption{\textcolor{black}{Tracking change in leg damping coefficient}}
  \label{fig:slip_sysid}
  \vspace{-20pt}
\end{figure}

\textcolor{black}{We also demonstrated our SysID method in a contact rich environment using a robot hopper controlled by the SLIP model as described in \cite{raibert_slip}. The hopper consists of a 1kg body and a spring-loaded foot with 0.01 kg. The task is to hop on a flight of stairs successfully while swiftly adapting to the change of damping coefficient of the leg. Estimating damping coefficient accurately is crucial for the control policy to maintain sufficient level of energy in the system. Figure \ref{fig:slip_sysid} shows that our method tracks the change of damping coefficient accurately while the NAIVE benchmark fails to re-estimate the damping coefficient, leading to falling on the stairs (See the supplementary video).}

\subsection{Experiment on real robot}
\begin{figure}[t]
  \centering
  \includegraphics[width=\linewidth]{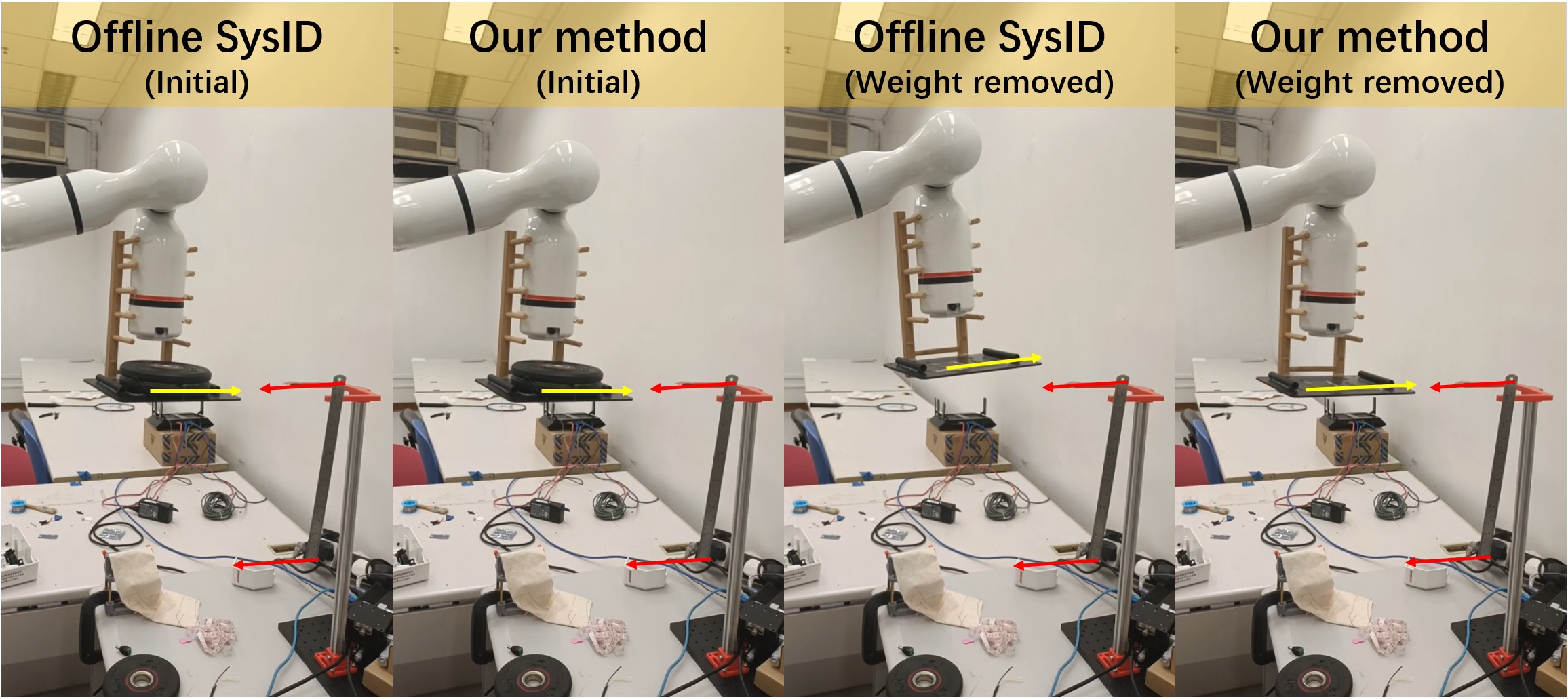}
  \vspace{-15pt}
  \caption{Experiments on a physical robot. The task is to control the end-effector moving between the two red arrows.}
  \label{fig:real_arm}
  \vspace{-20pt}
\end{figure}

We validated the effectiveness of online SysID on a physical 7-dof robot arm, Rokae Xmate3 Pro. The task is to control the end-effector to move vertically following a sinusoidal trajectory using PD controllers, under the online changes of end-effector payload mass as well as changing mass attach to a body link. To maintain precise control, we apply gravity compensation, which requires an accurate model of the robot to be available. We compared our method to an offline SysID method in which the initial mass of end-effector mass is correctly identified and applied, but does not adapt to the change of mass during the task execution. Figure \ref{fig:real_arm} and the attached video show that our method is able to swiftly identify the new mass after adding and removing payload from the robot thus maintaining precise control throughout ($2$ cm tracking error). In contrast, the offline SysID was not able to reach the targets precisely after the payload was removed ($7$ cm tracking error).


%% file: conclusion.tex
\section{Conclusion and Limitation} 
\label{sec:conclusion}
We developed a framework that performs online motion planning and system identification simultaneously in real-time using differentiable physics simulation. Our online system identification algorithm is robust against noisy  observations and is able to detect changes in the environments in real-time. Our model predictive control adaptively adjusts the optimizing window to ensure effective real-time control.

Optimizing complex trajectories with contacts remains challenging for our method. While in principle our method can be extended to scenarios with contact, in practice the convergence of the iLQR planner varies widely both in terms of the computation time and the quality of solution due to ambiguous contact gradient and rough loss landscape. A more effective controller is a critical next step for our method. 


%% file: root.bbl
\begin{thebibliography}{10}
\providecommand{\url}[1]{#1}
\csname url@samestyle\endcsname
\providecommand{\newblock}{\relax}
\providecommand{\bibinfo}[2]{#2}
\providecommand{\BIBentrySTDinterwordspacing}{\spaceskip=0pt\relax}
\providecommand{\BIBentryALTinterwordstretchfactor}{4}
\providecommand{\BIBentryALTinterwordspacing}{\spaceskip=\fontdimen2\font plus
\BIBentryALTinterwordstretchfactor\fontdimen3\font minus
  \fontdimen4\font\relax}
\providecommand{\BIBforeignlanguage}[2]{{%
\expandafter\ifx\csname l@#1\endcsname\relax
\typeout{** WARNING: IEEEtran.bst: No hyphenation pattern has been}%
\typeout{** loaded for the language `#1'. Using the pattern for}%
\typeout{** the default language instead.}%
\else
\language=\csname l@#1\endcsname
\fi
#2}}
\providecommand{\BIBdecl}{\relax}
\BIBdecl

\bibitem{sim2real}
P.~F. Christiano, Z.~Shah, I.~Mordatch, J.~Schneider, T.~Blackwell, J.~Tobin,
  P.~Abbeel, and W.~Zaremba, ``Transfer from simulation to real world through
  learning deep inverse dynamics model,'' \emph{CoRR}, 2016.

\bibitem{sysid_survey}
K.~J. {\AA}str{\"o}m and P.~Eykhoff, ``System identification—a survey,''
  \emph{Automatica}, 1971.

\bibitem{RLS}
M.~Kaess, A.~Ranganathan, and F.~Dellaert, ``isam: Incremental smoothing and
  mapping,'' \emph{{IEEE} TR-O}, 2008.

\bibitem{DOOMED}
N.~D. Ratliff, F.~Meier, D.~Kappler, and S.~Schaal, ``{DOOMED:} direct online
  optimization of modeling errors in dynamics,'' \emph{Big Data}, 2016.

\bibitem{NLRLS}
F.~Ding, X.~Wang, Q.~Chen, and Y.~Xiao, ``Recursive least squares parameter
  estimation for a class of output nonlinear systems based on the model
  decomposition,'' \emph{CSSP}, 2016.

\bibitem{DiffCloth}
J.~Liang, M.~C. Lin, and V.~Koltun, ``Differentiable cloth simulation for
  inverse problems,'' in \emph{NeurIPS}, 2019.

\bibitem{LCP-Physics}
F.~de~Avila~Belbute{-}Peres, K.~A. Smith, K.~R. Allen, J.~Tenenbaum, and J.~Z.
  Kolter, ``End-to-end differentiable physics for learning and control,'' in
  \emph{NeurIPS}, 2018.

\bibitem{GEO}
T.~Lee, P.~M. Wensing, and F.~C. Park, ``Geometric robot dynamic
  identification: {A} convex programming approach,'' \emph{TR-O}, 2020.

\bibitem{DLN}
M.~Lutter, C.~Ritter, and J.~Peters, ``Deep lagrangian networks: Using physics
  as model prior for deep learning,'' in \emph{ICLR}, 2019.

\bibitem{DUST}
L.~Barcelos, A.~Lambert, R.~Oliveira, P.~Borges, B.~Boots, and F.~Ramos, ``Dual
  online stein variational inference for control and dynamics,'' in \emph{RSS},
  2021.

\bibitem{EMPPI}
I.~Abraham, A.~Handa, N.~D. Ratliff, K.~Lowrey, T.~D. Murphey, and D.~Fox,
  ``Model-based generalization under parameter uncertainty using path integral
  control,'' \emph{{IEEE} RA-L}, 2020.

\bibitem{UP-OSI}
W.~Yu, J.~Tan, C.~K. Liu, and G.~Turk, ``Preparing for the unknown: Learning a
  universal policy with online system identification,'' in \emph{RSS}, 2017.

\bibitem{SimGAN}
Y.~Jiang, T.~Zhang, D.~Ho, Y.~Bai, C.~K. Liu, S.~Levine, and J.~Tan, ``Simgan:
  Hybrid simulator identification for domain adaptation via adversarial
  reinforcement learning,'' in \emph{ICRA}, 2021.

\bibitem{GPSysID}
J.~Boedecker, J.~T. Springenberg, J.~W{\"{u}}lfing, and M.~A. Riedmiller,
  ``Approximate real-time optimal control based on sparse gaussian process
  models,'' in \emph{{IEEE} ADPRL}, 2014.

\bibitem{CommonLyapunov}
L.~Vu and D.~Liberzon, ``Common lyapunov functions for families of commuting
  nonlinear systems,'' \emph{SC-L}, 2005.

\bibitem{DBLP:conf/rss/TanZCIBHBV18}
J.~Tan, T.~Zhang, E.~Coumans, A.~Iscen, Y.~Bai, D.~Hafner, S.~Bohez, and
  V.~Vanhoucke, ``Sim-to-real: Learning agile locomotion for quadruped
  robots,'' in \emph{RSS}, 2018.

\bibitem{DBLP:conf/icra/PengAZA18}
X.~B. Peng, M.~Andrychowicz, W.~Zaremba, and P.~Abbeel, ``Sim-to-real transfer
  of robotic control with dynamics randomization,'' in \emph{ICRA}, 2018.

\bibitem{DBLP:journals/corr/abs-1901-08652}
J.~Hwangbo, J.~Lee, A.~Dosovitskiy, D.~Bellicoso, V.~Tsounis, V.~Koltun, and
  M.~Hutter, ``Learning agile and dynamic motor skills for legged robots,''
  \emph{CoRR}, 2019.

\bibitem{DBLP:journals/corr/abs-2011-01891}
I.~Exarchos, Y.~Jiang, W.~Yu, and C.~K. Liu, ``Policy transfer via kinematic
  domain randomization and adaptation,'' in \emph{ICRA}, 2021.

\bibitem{RMA}
A.~Kumar, Z.~Fu, D.~Pathak, and J.~Malik, ``{RMA:} rapid motor adaptation for
  legged robots,'' in \emph{RSS}, 2021.

\bibitem{degrave2019differentiable}
J.~Degrave, M.~Hermans, J.~Dambre, and F.~Wyffels, ``A differentiable physics
  engine for deep learning in robotics,'' \emph{Frontiers Neurorobotics}, 2019.

\bibitem{difftaichi}
Y.~Hu, L.~Anderson, T.~Li, Q.~Sun, N.~Carr, J.~Ragan{-}Kelley, and F.~Durand,
  ``Difftaichi: Differentiable programming for physical simulation,'' in
  \emph{ICLR}, 2020.

\bibitem{Zhong2021NIPS}
Y.~D. Zhong, B.~Dey, and A.~Chakraborty, ``Extending lagrangian and hamiltonian
  neural networks with differentiable contact models,'' in \emph{NeurIPS},
  2021.

\bibitem{DiSECT}
E.~Heiden, M.~Macklin, Y.~S. Narang, D.~Fox, A.~Garg, and F.~Ramos, ``Disect:
  {A} differentiable simulation engine for autonomous robotic cutting,'' in
  \emph{RSS}, 2021.

\bibitem{Wang2021IROS}
K.~Wang, M.~Aanjaneya, and K.~E. Bekris, ``Sim2sim evaluation of a novel
  data-efficient differentiable physics engine for tensegrity robots,'' in
  \emph{IROS}, 2021.

\bibitem{Lidec2021ICRA}
Q.~Le~Lidec, I.~Kalevatykh, I.~Laptev, C.~Schmid, and J.~Carpentier,
  ``Differentiable simulation for physical system identification,'' \emph{IEEE
  RA-L}, 2021.

\bibitem{geilinger2020add}
M.~Geilinger, D.~Hahn, J.~Zehnder, M.~B{\"{a}}cher, B.~Thomaszewski, and
  S.~Coros, ``{ADD:} analytically differentiable dynamics for multi-body
  systems with frictional contact,'' \emph{{ACM} ToG}, 2020.

\bibitem{Xu2021RSS}
J.~Xu, T.~Chen, L.~Zlokapa, M.~Foshey, W.~Matusik, S.~Sueda, and P.~Agrawal,
  ``An end-to-end differentiable framework for contact-aware robot design,'' in
  \emph{RSS}, 2021.

\bibitem{Lutter2021ICRA}
M.~Lutter, J.~Silberbauer, J.~Watson, and J.~Peters, ``A differentiable newton
  euler algorithm for multi-body model learning,'' \emph{ICRA}, 2020.

\bibitem{heiden2020augmenting}
E.~Heiden, D.~Millard, E.~Coumans, Y.~Sheng, and G.~S. Sukhatme, ``Neuralsim:
  Augmenting differentiable simulators with neural networks,'' in \emph{ICRA},
  2021.

\bibitem{ContactNets}
S.~Pfrommer, M.~Halm, and M.~Posa, ``Contactnets: Learning discontinuous
  contact dynamics with smooth, implicit representations,'' in \emph{CoRL},
  2020.

\bibitem{nimble}
K.~Werling, D.~Omens, J.~Lee, I.~Exarchos, and C.~K. Liu, ``Fast and
  feature-complete differentiable physics engine for articulated rigid bodies
  with contact constraints,'' in \emph{RSS}, 2021.

\bibitem{Zamora2021ICML}
M.~Zamora, M.~Peychev, S.~Ha, M.~T. Vechev, and S.~Coros, ``{PODS:} policy
  optimization via differentiable simulation,'' in \emph{ICML}, 2021.

\bibitem{persistent}
K.~S. Narendra and A.~M. Annaswamy, ``Persistent excitation in adaptive
  systems,'' \emph{International Journal of Control}, 1987.

\bibitem{IPOPT}
A.~W{\"a}chter and L.~T. Biegler, ``On the implementation of an interior-point
  filter line-search algorithm for large-scale nonlinear programming,''
  \emph{Mathematical Programming}, 2006.

\bibitem{iLQR}
W.~Li and E.~Todorov, ``Iterative linear quadratic regulator design for
  nonlinear biological movement systems,'' in \emph{ICINCO}, 2004.

\bibitem{raibert_slip}
M.~Raibert, ``Hopping in legged systems: Modeling and simulation for the
  two-dimensional one-legged case,'' \emph{TSMC}, 1984.

\end{thebibliography}
